\newif\iffinalcopy
\finalcopyfalse
\def\finalcopy{\global\finalcopytrue}
\finalcopy

\documentclass[runningheads]{llncs}
\usepackage{graphicx}
\usepackage{amsmath,amssymb} 
\iffinalcopy\else\usepackage{ruler}\fi
\usepackage{color}
\usepackage{xspace}
\usepackage[width=122mm,left=12mm,paperwidth=146mm,height=193mm,top=12mm,paperheight=217mm]{geometry}

\makeatletter
\DeclareRobustCommand\onedot{\futurelet\@let@token\@onedot}
\def\@onedot{\ifx\@let@token.\else.\null\fi\xspace}

\def\eg{\emph{e.g}\onedot} 
\def\ie{\emph{i.e}\onedot} 
\def\cf{\emph{c.f}\onedot} 
 
\def\wrt{w.r.t\onedot} 
\def\etal{\emph{et al}\onedot}
\makeatother

\usepackage{subcaption}
\begin{document}
\pagestyle{headings}
\mainmatter
\def\ECCV18SubNumber{2874}  

\newcommand\todoeric[1]{\textbf{TODO: }\textcolor{red}{#1}}
\newcommand*{\resp}{resp.\@\xspace}
\newcommand*{\ff}{et seqq.\@\xspace}


\title{iPose: Instance-Aware 6D Pose Estimation of Partly Occluded Objects} 

\iffinalcopy
\titlerunning{iPose: Instance-Aware 6D Pose Estimation of Partly Occluded Objects}
\authorrunning{Hosseini Jafari, Mustikovela, Pertsch, Brachmann, Rother}
\author{Omid Hosseini Jafari\thanks{Equal contribution}, Siva Karthik Mustikovela$^{\star}$\\Karl Pertsch, Eric Brachmann, Carsten Rother}
\institute{Visual Learning Lab - Heidelberg University (HCI/IWR)\\ \url{http://vislearn.de}}
\else
\titlerunning{ECCV-18 submission ID \ECCV18SubNumber}
\authorrunning{ECCV-18 submission ID \ECCV18SubNumber}
\author{Anonymous ECCV submission}
\institute{Paper ID \ECCV18SubNumber}
\fi

\maketitle

\begin{abstract}
We address the task of 6D pose estimation of known rigid objects from single input images in scenarios where the objects are partly occluded.
Recent RGB-D-based methods are robust to moderate degrees of occlusion.
For RGB inputs, no previous method works well for partly occluded objects.
Our main contribution is to present the first deep learning-based system that estimates accurate poses for partly occluded objects from RGB-D and RGB input. 
We achieve this with a new instance-aware pipeline that decomposes 6D object pose estimation into a sequence of simpler steps, where each step removes specific aspects of the problem. 
The first step localizes all known objects in the image using an instance segmentation network, and hence eliminates surrounding clutter and occluders. 
The second step densely maps pixels to 3D object surface positions, so called object coordinates, using an encoder-decoder network, and hence eliminates object appearance. 
The third, and final, step predicts the 6D pose using geometric optimization. 
We demonstrate that we significantly outperform the state-of-the-art for pose estimation of partly occluded objects for both RGB and RGB-D input.
\end{abstract}


\section{Introduction}

Localization of object instances from single input images has been a long-standing goal in computer vision. 
The task evolved from simple 2D detection to full 6D pose estimation, \ie estimating the 3D position and 3D orientation of the object relative to the observing camera. 
Early approaches relied on objects having sufficient texture to match feature points \cite{lowe01local}.
Later, with the advent of consumer depth cameras \cite{kinect}, research focused on texture-less objects \cite{hinterstoisser2011linemod} in increasingly cluttered environments.
Today, heavy occlusion of objects is the main performance benchmark for one-shot pose estimation methods.
Object occlusion occurs in all scenarios, apart from artificial settings, hence robustness to occlusion is crucial in applications like augmented reality or robotics.

Recent RGB-D-based methods \cite{michel2017,Hinterstoisser2016} are robust to moderate degrees of object occlusion.
However, depth cameras fail under certain conditions, \eg with intense sunlight, and RGB cameras are prevalent on many types of devices.
Hence, RGB-based methods still have high practical relevance.
In this work, we present a system for 6D pose estimation of rigid object instances from single input images. 
The system performs well for partly occluded objects. That means for both input modalities, RGB-D and RGB, it clearly outperforms the accuracy of previous methods.

During the last decade, computer vision has seen a large shift towards learning-based methods.
In particular, \emph{deep learning}, \ie training multi-layered neural networks, has massively improved accuracy and robustness for many tasks, most notably object recognition \cite{alexnet2012}, object detection \cite{girshick15fastrcnn,yolo2016,ssd2016} and semantic segmentation \cite{fcn2015,Dai2016CVPR,he2017iccv}.
While 6D object pose estimation has also benefited from deep learning to some extent, with recent methods being able to estimate accurate poses in real time from single RGB images \cite{bb82017,kehl2017iccv,Tekin18cvpr}, the same does not hold when objects are partly occluded.
In this case, aforementioned methods, despite being trained with partly occluded objects, either break down \cite{kehl2017iccv,Tekin18cvpr} or have to simplify the task by estimating poses from tight crops around the ground truth object position \cite{bb82017}.
To the best of our knowledge, we are the first to show that deep learning can improve results considerably for objects that are moderately to heavily occluded, particularly for the difficult case of RGB input.

\begin{figure*}[t!]
\centering
  \includegraphics[width=\linewidth]{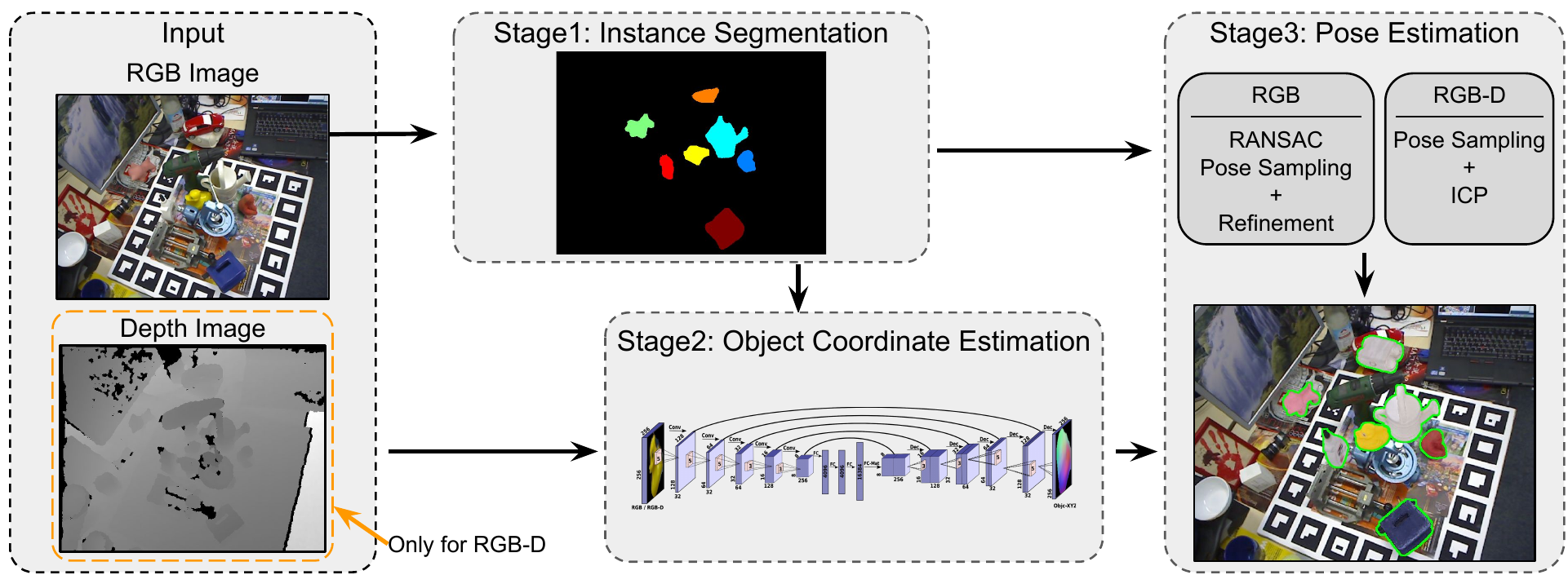}
  \label{fig:boat1}
   \caption{\textbf{Illustration of our modular, 3-stage pipeline} for both RGB and RGB-D input images.}
\label{fig::pipeline}
\vspace{-0.5cm}
\end{figure*}

At the core, our method decomposes the 6D pose estimation problem into a sequence of three sub-tasks, or modules (see Fig.~\ref{fig::pipeline}).
We first detect the object in 2D, then we locally regress correspondences to the 3D object surface, and, finally, we estimate the 6D pose of the object.
With each sub-task, we can remove specific aspects of the problem, such as object background and object appearance.
In the first module, 2D detection is implemented by an instance segmentation network which estimates a tight mask for each object.
Thus, we can separate the object from surrounding clutter and occluders, making the following steps invariant to the object environment, and allowing us to process each detected instance individually.
In the second module, we present an encoder-decoder architecture for densely regressing so-called \emph{object coordinates} \cite{brachmann2014pose6d}, \ie 3D points in the local coordinate frame of the object which define 2D-3D correspondences between the image and the object.
The third module is a purely geometric pose optimization which is not learned from data because all aspects of object appearance have been removed in the previous steps.
Since we estimate 6D poses successively from 2D instance segmentation, we call our approach \emph{iPose}, short for ``instance-aware pose estimation''.

Our decomposition strategy is conceptually simple, but we show that it is considerably superior to other deep learning-based methods that try to reason about different aspects of these steps jointly.
In particular, several recent works propose to extend state-of-the-art object detection networks to output 6D object poses directly. 
Kehl \etal \cite{kehl2017iccv} extend the SSD object detector \cite{ssd2016} to recognize discretized view-points of specific objects, \ie re-formulating pose regression as a classification problem.
Similarly, Tekin \etal \cite{Tekin18cvpr} extend the YOLO object detector \cite{yolo2016} by letting image grid cells predict object presence, and simultaneously the 6D pose.
Both approaches are highly sensitive to object occlusion, as we will show in the experimental evaluation. 
Directly predicting the 6D pose from observed object appearance is challenging, due to limited training data and innumerable occlusion possibilities.

We see three reasons for the success of our approach.
Firstly, we exploit the massive progress in object detection and instance segmentation achieved by methods like MNC \cite{Dai2016CVPR} and Mask R-CNN \cite{he2017iccv}. This is similar in spirit to the work of \cite{kehl2017iccv,Tekin18cvpr}, but instead of extending the instance segmentation to predict 6D poses directly, 
we use it as a decoupled component within our step-by-step strategy.
Secondly, the rich structural output of our dense object coordinate regression step allows for a geometric hypothesize-and-verify approach that can yield a good pose estimate even if parts of the prediction are incorrect, \eg due to occlusion. 
Such a robust geometry-based step is missing in previous deep learning-based approaches \cite{bb82017,kehl2017iccv,Tekin18cvpr}.
Thirdly, we propose a new data augmentation scheme specifically designed for the task of 6D object pose estimation. 
Data augmentation is a common aspect of learning-based pose estimation methods, since training data is usually scarce.
Previous works have placed objects at random 2D locations over arbitrary background images \cite{brachmann2016,bb82017,kehl2017iccv}, which yields constellations where objects occlude each other in physically impossible ways.
In contrast, our data augmentation scheme infers a common ground plane from ground truth poses and places additional objects in a physically plausible fashion. 
Hence, our data augmentation results in more realistic occlusion patterns which we found crucial for obtaining good results.

\noindent We summarize our main {\bf contributions}:
\begin{itemize}
\item We propose \emph{iPose}, a new deep learning architecture for 6D object pose estimation which is remarkably robust \wrt object occlusion, using a new three-step task decomposition approach.
\item We are the first to surpass the state-of-the-art for partly occluded objects with a deep learning-based approach for both RGB-D and RGB inputs.
\item We present a new data augmentation scheme for object pose estimation which generates physically plausible occlusion patterns, crucial for obtaining good results.

\end{itemize}
\section{Related Work}

Below, we give an overview of previous methods for 6D object pose estimation.
Note that there is a body of work regarding pose estimation of object categories, specifically in the context of autonomous driving on datasets like KITTI \cite{kitti2012CVPR}, see \eg \cite{3dprop2015,chen2016monocular,mulauto2017,manta2017}. 
Because of intra-class variability, these approaches often estimate coarse viewpoints or constrained poses, \eg 3D poses on a ground plane.
In this work, we consider the different task of estimating full 6D poses of specific, rigid object instances.

Early pose estimation methods were based on matching sparse features \cite{lowe01local} or templates \cite{Huttenlocher93comparingimages}.
Templates work well for texture-less objects where sparse feature detectors fail to identify salient points.
Hinterstoisser \etal proposed the LINEMOD templates \cite{hinterstoisser2011linemod}, which combine gradient and normal cues for robust object detection given RGB-D inputs.
Annotating the template database with viewpoint information facilitates accurate 6D pose estimation \cite{Hinterstoisser12model,rios2013discriminatively,hodan2015,hashmod2016,Konishi2016}.
An RGB version of LINEMOD \cite{hinterstoisser2012pami} is less suited for pose estimation \cite{brachmann2016}.
In general, template-based methods suffer from sensitivity to occlusion \cite{brachmann2014pose6d}.

With a depth channel available, good results have been achieved by voting-based schemes \cite{alykhan2014latenthough,zach2015dynamic,Doumanoglou2016,Kehl2016,Drost10model,Hinterstoisser2016}.
In particular, Drost \etal \cite{Drost10model} cast votes by matching point-pair features which combine normal and distance information. 
Recently, the method was considerably improved in \cite{Hinterstoisser2016} by a suitable sampling scheme, resulting in a purely geometric method that achieves state-of-the-art results for partly occluded objects given RGB-D inputs.
Our deep learning-based pipeline achieves higher accuracy, and can also be applied to RGB images.

Recently, deep learning-based methods have become increasingly popular for object pose estimation from RGB images.
Rad and Lepetit \cite{bb82017} presented the BB8 pipeline which resembles our decomposition philosophy to some extent.
However, their processing steps are more tightly coupled. 
For example, their initial detection stage does not segment the object, and can thus not remove object background.
Also, they regress the 6D pose by estimating the 2D location of a sparse set of control points.
We show that dense 3D object coordinate regression provides a richer output which is essential for robust geometric pose optimization.
Rad and Lepetit \cite{bb82017} evaluate BB8 on occluded objects but restrict pose prediction to image crops around the ground truth object position\footnote{Their experimental setup which relies on ground truth crops is not explicitly described in \cite{bb82017}, but we verified this information from a private email exchange with the authors of \cite{bb82017}.}.
Our approach yields superior results for partly occluded objects \emph{without} using prior knowledge about object position.

Direct regression of a 6D pose vector by a neural network, \eg proposed by Kendall \etal for camera localization \cite{kendall2015convolutional}, exhibits low accuracy \cite{brachmann2017dsac}.
The works discussed in the introduction, \ie Kehl \etal \cite{kehl2017iccv} and Tekin \etal \cite{Tekin18cvpr}, also regress object pose directly but make use of alternative pose parametrizations, namely discrete view point classification \cite{kehl2017iccv}, or sparse control point regression \cite{Tekin18cvpr} similar to BB8 \cite{bb82017}.
We do {\it not} predict the 6D pose directly, but follow a step-by-step strategy to robustly obtain the 6D pose despite strong occlusions. 

Object coordinates have been used previously for object pose estimation from RGB-D \cite{brachmann2014pose6d,krull2015analysis,michel2017} or RGB inputs \cite{brachmann2016}. 
In these works, random forest matches image patches to 3D points in the local coordinate frame of the object, and the pose is recovered by robust, geometric optimization.
Because few correct correspondences suffice for a pose estimate, these methods are inherently robust to object occlusion. 
In contrast to our work, they combine object coordinate prediction and object segmentation in a single module, using random forests. 
These two tasks are disentangled in our approach, with the clear advantage that each individual object mask is known for object coordinate regression. 
In this context, we are also the first to successfully train a neural network for object coordinate regression of known objects. 
Overall, we report superior pose accuracy for partly occluded objects using RGB and RGB-D inputs.
Note that recently Behl \etal \cite{Behl2017ICCV} have trained a network for object coordinate regression of vehicles (\ie object class). 
However, our network, training procedure, and data augmentation scheme differ from \cite{Behl2017ICCV}.


%

To cope well with limited training data, we propose a new data augmentation scheme which generates physically plausible occlusion patterns. 
While plausible data augmentation is becoming common in object class detection works, see \eg \cite{virtkitti2016,ardriving2017,occaware2017}, our scheme is tailored specifically towards object instance pose estimation where previous works resorted to pasting 2D object crops on arbitrary RGB backgrounds \cite{brachmann2016,bb82017,kehl2017iccv}.
We found physically plausible data augmentation to be crucial for obtaining good results for partly occluded objects.

To summarize, only few previous works have addressed the challenging task of pose estimation of partly occluded objects from single RGB or RGB-D inputs. 
We present the first viable deep learning approach for this scenario, improving state-of-the-art accuracy considerably for both input types.

\section{Method}
In this section, we describe our three-stage, instance-aware approach for 6D object pose estimation. 
The overall workflow of our method is illustrated in Fig.~\ref{fig::pipeline}. 
Firstly, we obtain all object instances in a given image using an instance segmentation network (Sec.~\ref{subsec::ins_seg}). 
Secondly, we estimate dense 3D object coordinates for each instance using an encoder-decoder network (Sec.~\ref{subsec::objc_est}). 
Thirdly, we use the pixel-wise correspondences between predicted object coordinates and the input image to sample 6D pose hypotheses, and further refine them using an iterative geometric optimization (Sec.~\ref{subsec::pose_est}). 
In Sec.~\ref{subsec::data_aug}, we describe our object-centric data augmentation procedure which we use to generate additional training data with realistic occlusions for the encoder-decoder network of step 2.

We denote the RGB input to our pipeline as \textit{I} and RGB-D input as \textit{I-D}.
$\mathcal{K}=\{1,...,K\}$ is a set of all known object classes, a subset of which could be present in the image. 
The goal of our method is to take an image \textit{I}/\textit{I-D} containing $n$ objects $\mathcal{O}=\{O_1, ...,O_n\}$, each of which has a class from $\mathcal{K}$, and to estimate their 6D poses. 
Below, we describe each step of our pipeline in detail.

\subsection{Stage 1: Instance Segmentation}\label{subsec::ins_seg}
The first step of our approach, instance segmentation, recognizes the identity of each object, and produces a fine grained mask.
Thus we can separate the \mbox{RGB(-D)} information pertaining only to a specific object from surrounding clutter and occluders. 
To achieve this, we utilize instance segmentation frameworks such as  \cite{Dai2016CVPR,he2017iccv}.  
Given an input \textit{I}, the output of this network is a set of n instance masks $\mathcal{M}=\{M_1,...,M_n\}$ and an object class $k\in\mathcal{K}$ for each mask. 


\subsection{Stage 2: Object Coordinate Regression}\label{subsec::objc_est}
An object coordinate denotes the 3D position of an object surface point in the object's local coordinate frame.
Thus given a pixel location $p$ and its predicted object coordinate $C$, a $(p,C)$ pair defines a correspondence between an image $I$ and object $O$. 
Multiple such correspondences, at least three for RGB-D data and four for RGB data, are required to recover the 6D object pose (see Sec.~\ref{subsec::pose_est}).
In order to regress pixelwise object coordinates $C$ for each detected object, we use a CNN with an encoder-decoder style architecture with skip connections. 
The encoder consists of 5 convolutional layers with a stride of 2 in each layer, followed by a set of 3 fully connected layers. 
The decoder has 5 deconvolutional layers followed by the 3 layer output corresponding to 3-dimensional object coordinates. 
Skip connections exist between symmetrically opposite conv-deconv layers. 
As input for this network, we crop a detected object using its estimated mask $M$, resize and pad the crop to a fixed size, and pass it through the object coordinate network. 
The output of this network has 3 channels containing the pixelwise X, Y and Z values of object coordinates $C$ for mask $M$.
We train separate networks for RGB and RGB-D inputs.

\subsection{Stage 3: Pose Estimation}\label{subsec::pose_est}
In this section, we describe the geometric pose optimization step of our approach for RGB-D and RGB inputs, respectively. 
This step is not learned from data, but recovers the 6D object pose from the instance mask $M$ of stage 1 and the object coordinates $C$ of stage 2.\\

{\bf RGB-D Setup.}
Our pose estimation process is inspired by the original object coordinate framework of \cite{brachmann2014pose6d}.
Compared to \cite{brachmann2014pose6d}, we use a simplified scoring function to rank pose hypotheses, and an Iterative Closest Point (ICP) refinement.

In detail, we use the depth channel and the mask $M_{O}$ to calculate a 3D point cloud $P_{O}$ associated with object $O$ \wrt the coordinate frame of the camera.
Also, stage 2 yields the pixelwise predicted object coordinates $C_{O}$. 
We seek the 6D pose $H^*_{O}$ which relates object coordinates $C_{O}$ with the point cloud $P_{O}$.
For ease of notation, we drop the subscript $O$, assuming that we are describing the process for that particular object instance. 
We randomly sample three pixels $j_1, j_2, j_3$ from mask $M$, from which we establish three 3D-3D correspondences $(P^{j_1}, C^{j_1})$, $(P^{j_2}, C^{j_2})$, $(P^{j_3}, C^{j_3})$. 
We use the Kabsch algorithm \cite{kabsch1976} to compute the pose hypothesis $H_i$ from these correspondences. 
Using $H_i$, we transform $C^{j_{1}}$, $C^{j_{2}}$, $C^{j_{3}}$ from the object coordinate frame to the camera coordinate frame. 
Let these transformed points be $T^j$. 
We compute the Euclidean distance, $\lVert P^{j}, T^{j}\rVert$, and 
if the distances of all three points are less than 10\% of the object diameter, we add $H_i$ to our hypothesis pool. 
We repeat this process until we have collected 210 hypotheses. 
For each hypothesis $H$, we obtain a point cloud $P^{*}(H)$ in the camera coordinate system via rendering the object CAD model. 
This lets us score each hypothesis using
\begin{equation}
S_{\text{RGB-D}}(H) = \frac{\sum_{j\in M}\left[||P^{j} - P^{*j}(H)|| < d/10\right]}{|M|},
\end{equation}
where $[\cdot]$ returns 1 if the enclosed condition is true, and the sum is over pixels inside the mask $M$ and normalized.
The score $S_{\text{RGB-D}}(H)$ computes the average number the pixels inside the mask for which the rendered camera coordinates $P^{*j}(H)$ and the observed camera coordinates $P^{j}$ agree, up to a tolerance of 10\% of the object diameter $d$. 
From the initial pool of 210 hypotheses we select the top 20 according to the score $S_{\text{RGB-D}}(H)$.
Finally, for each selected hypothesis, we perform ICP refinement with $P$ as the target, the CAD model vertices as the source, and $H_{i}$ as initialization. 
We choose the pose with the lowest ICP fitting error $H_{\text{ICP}}$ for further refinement. 

\noindent \textbf{Rendering-Based Refinement.} 
Under the assumption that the estimate $H_{\text{ICP}}$ is already quite accurate, and using the instance mask $M$, we perform the following additional refinement:
using $H_{\text{ICP}}$, we render the CAD model to obtain a point cloud $P_r$ of the visible object surface.
This is in contrast to the previous ICP refinement where all CAD model vertices were used. 
We fit $P_r$ inside the mask $M$ to the observed point cloud $P$ via ICP, to obtain a refining transformation $H_{\text{ref}}$. 
This additional step pushes $P_r$ towards the observed point cloud $P$, providing a further refinement to $H_{\text{ICP}}$. 
The final pose is thus obtained by $H^*_{\text{RGB-D}} = H_{\text{ICP}}*H_{\text{ref}}$.

Our instance-based approach is a clear advantage in both refinement steps, since we can use the estimated mask to precisely carve out the observed point cloud for ICP.

\begin{figure*}
\centering
  \includegraphics[width=0.9\linewidth]{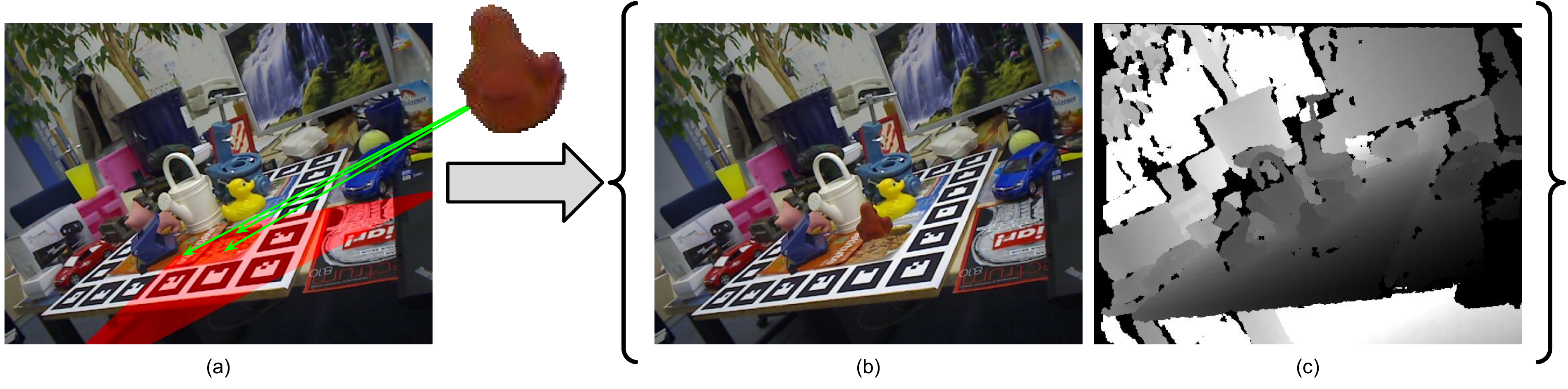}
  \caption{\textbf{Object centric data augmentation pipeline.} (a) If the cropped object (Ape) is inserted within the red area, it can cause a physically plausible occlusion for the center object (Can). (b) shows the resulting augmented RGB image, and (c) shows the resulting augmented depth image.}
  \label{fig::advanced_data_aug_pipeline}
  \vspace{-0.5cm}
\end{figure*}

{\bf RGB Setup.}
Given RGB data, we follow Brachmann \etal \cite{brachmann2016} and estimate the pose of the objects through hypotheses sampling \cite{brachmann2014pose6d} and pre-emptive RANSAC \cite{DBLP:conf/cvpr/ShottonGZICF13}. 
At this stage, the predicted object mask $M$ and the predicted object coordinates $C$ inside the mask are available. 
For each pixel $j$ at the 2D position $p_j$ inside $M$, the object coordinate network estimates a 3D point $C^j$ in the local object coordinate system. 
Thus, we can sample 2D-3D correspondences between 2D points of the image and 3D object coordinate points from the area inside the object mask. 
Our goal is to search for a pose hypothesis $H^*$ which maximizes the following score:
\begin{equation}
S_{\text{RGB}}(H) = \sum_{j\in M} \left[\lVert p_j-AHC^j\rVert_2<\tau_\textrm{in} \right] ,
\end{equation}
where $A$ is the camera projection matrix, $\tau_\textrm{in}$ is a threshold, and $[\cdot]$ is $1$ if the statement inside the bracket is true, otherwise $0$. 
The score $S_{\text{RGB}}(H)$ counts the number of pixel-residuals of re-projected object coordinate estimates which are below $\tau_\textrm{in}$. 
We use pre-emptive RANSAC to maximize this objective function.
We start by drawing four correspondences from the predicted mask $M$. 
Then, we solve the perspective-n-point problem (PnP) \cite{gao2003complete,lepetit2009epnp} to obtain a pose hypothesis. 
If the re-projection error of the initial four correspondences is below threshold $\tau_\textrm{in}$ we keep the hypothesis. 
We repeat this process until $256$ pose hypotheses have been collected.
We score each hypothesis with $S_{\text{RGB}}(H)$, but only using a sub-sampling of $N$ pixels inside the mask for faster computation.
We sort the hypotheses by score and discard the lower half. 
We refine the remaining hypotheses by re-solving PnP using their inlier pixels according to $S_{\text{RGB}}(H)$. 
We repeat scoring with an increased pixel count $N$, discarding and refining hypotheses until only one hypothesis $H^*_{\text{RGB}}$ remains as the final estimated pose.

\vspace{-0.2cm}
\subsection{Data Augmentation}
\label{subsec::data_aug}
Data augmentation is crucial for creating the amount of data necessary to train a deep neural network. 
Additionally, data augmentation can help to reduce dataset bias, and introduce novel examples for the network to train on.
One possibility for data augmentation is to paste objects on a random background, where mutually overlapping objects occlude each other.
This is done \eg in \cite{brachmann2016,bb82017,kehl2017iccv} and we found this strategy sufficient for training our instance segmentation network in step 1.
However, the resulting images and occlusion patterns are highly implausible, especially for RGB-D data where objects float in the scene, and occlude each other in physically impossible ways.
Training the object coordinate network in step 2 with such implausible data made it difficult for the network to converge and also introduced bias towards impossible object occlusion configurations. 
In the following, we present an object-centric data augmentation strategy which generates plausible object occlusion patterns, and analyze its impact on the dataset.
\begin{figure*}
\vspace{-0.0cm}
\centering
  \includegraphics[width=0.9\linewidth]{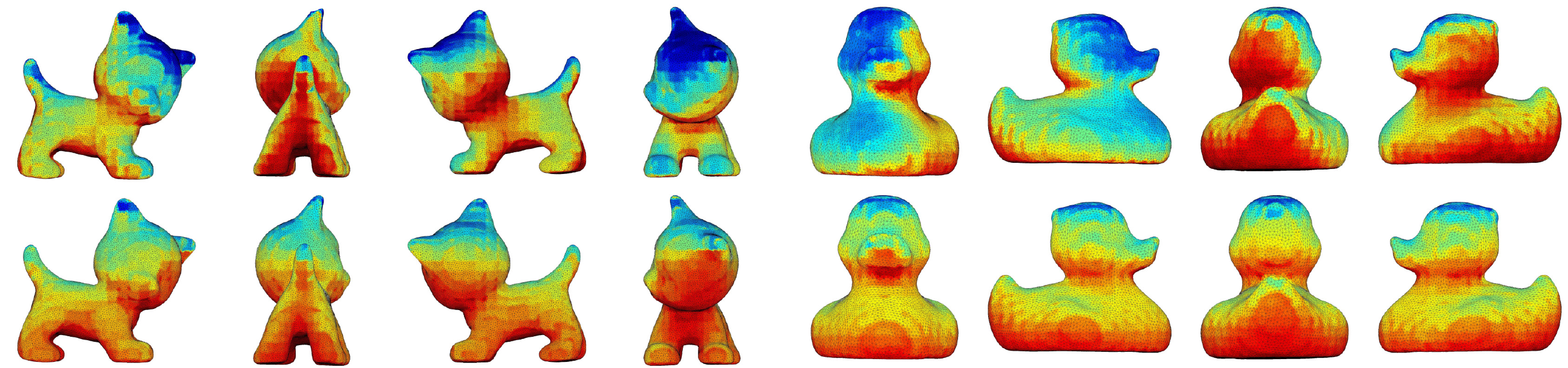}
  \caption{\textbf{Impact of our data augmentation.} The top row illustrates the on-object  occlusion distribution of the base training set before augmentation and the bottom row shows the same for augmented data using our object centric data augmentation. For a given part of the model, red indicates that the part is often occluded, while blue corresponds to rare occlusion in a given dataset.}
  \label{fig::on_object_occ} \vspace{-0.5cm}
\end{figure*}
%
We assume that for each target object $k$ in the set of all known objects $\mathcal{K}$, a sequence of images is available where the object is not occluded. 
For each image, we compute the ground plane on which the target object stands on, as well as the distance between its base point and the camera.
Then, as shown in Fig.~\ref{fig::advanced_data_aug_pipeline}(a)(red), a surface of interest is defined on the ground plane in front of the target object, representing a cone with an opening angle of $90^\circ$. 
Next, we search for images of other objects in $\mathcal{K}$, where the ground plane normal is close to that of the target object, and which are located in the defined surface of interest, based on their distance from the camera. 
Finally, by overlaying one or more of these chosen objects in front of the target object, we can generate multiple augmented RGB and depth images (\cf Fig.~\ref{fig::advanced_data_aug_pipeline}(b,c)).
Using this approach, the resulting occlusion looks physically correct for both the RGB and the depth image.


To analyze the impact of our data augmentation scheme, we visualize the distribution of partial occlusion on the object surface in the following way: 
we first discretize the 3D bounding box surrounding each object into $20\times20\times20$ voxels. 
Using the ground truth 6D pose and the 3D CAD model, we can render the full mask of the object.
Each pixel that lies inside the rendered mask but not inside the ground truth mask is occluded.
We can look-up the ground truth object coordinate of each occluded pixel, and furthermore the associated bounding box voxel. 
We use the voxels as histogram bins and visualize the occlusion frequency as colors on the surface of the 3D CAD model. 

The impact of our object-centric data augmentation for two objects of the LINEMOD dataset \cite{Hinterstoisser12model} is illustrated in Fig.~\ref{fig::on_object_occ}. 
Firstly, by looking at the visualization (top row), we notice that the un-augmented data contains biased occlusion samples (irregular distribution of blue and red patches) which could induce overfitting on certain object parts, leading to reduced performance of the object coordinate network of step 2. 
In the second row, we see that the augmented data has a more regular distribution of occlusion. 
This visualization reveals the bias in the base training set, and demonstrates the efficacy of our object-centric data augmentation procedure in creating unbiased training data samples.


\section{Experiments}
\vspace{-0.2cm}
In this section, we present various experiments quantifying the performance of our approach. 
In Sec.~\ref{subsec::dataset}, we introduce the dataset which we use for evaluating our system. 
In Sec.~\ref{subsec::sota_comparison}, we compare the performance of our approach to existing RGB and RGB-D-based pose estimation approaches. 
In Sec.~\ref{subsec::ablation_study}, we analyze the contribution of various modules of our approach to the final pose estimation performance. 
Finally, in Sec.~\ref{sub_sec::seg_eval} and \ref{sub_sec::objc_eval}, we discuss the performance of our instance segmentation and object coordinate estimation networks.
Please see the supplemental materials for a complete list of parameter settings of our pipeline.


\vspace{-0.3cm}
\subsection{Datasets and Implementation}\label{subsec::dataset}
We evaluate our approach on \textit{occludedLINEMOD}, a dataset published by Brachmann \etal \cite{brachmann2014pose6d}. 
It was created from the LINEMOD dataset \cite{Hinterstoisser12model} by annotating ground truth 6D poses for various objects in a sequence of 1214 RGB-D images. 
The objects are located on a table and embedded in dense clutter. 
Ground truth poses are provided for eight of these objects which, depending on the camera view, heavily occlude each other, making this dataset very challenging.
We test both our RGB and \mbox{RGB-D}-based methods on this dataset.
 
To train our system, we use a separate sequence from the LINEMOD dataset which was annotated by Michel \etal \cite{michel2017}. 
For ease of reference we call this the LINEMOD-M dataset.
LINEMOD-M comes with ground truth annotations of seven objects with mutual occlusion. 
One object of the test sequence, namely the Driller, is not present in this training sequence, so we do not report results for it. 
The training sequence is extremely limited in the amount of data it provides. 
Some objects are only seen from few viewpoints and with little occlusion, or occlusion affects only certain object parts. 

\noindent \textbf{Training Instance Segmentation.}
To train our instance segmentation network with a wide range of object viewpoints and diverse occlusion examples, we create synthetic images in the following way. 
We use RGB backgrounds from the NYUD dataset \cite{Silberman:ECCV12}, and randomly overlay them with objects picked from the original LINEMOD dataset \cite{Hinterstoisser12model}. 
While this data is physically implausible, we found it sufficient for training the instance segmentation component of our pipeline.
We combine these synthetic images with LINEMOD-M to obtain 9000 images with ground truth instance masks. 
We use Mask R-CNN \cite{he2017iccv} as our instance segmentation method.
For training, we use a learning rate of $1e$-$3$, momentum of $0.9$ and weight decay of $1e$-$4$. 
We initialize Mask R-CNN with weights trained on ImageNet \cite{imagenet_cvpr09}, and finetune on our training set.

\noindent \textbf{Training Object Coordinate Regression.}
For training the object coordinate estimation network, we found it important to utilize physically plausible data augmentation for best results. 
Therefore, we use the LINEMOD-M dataset along with the data obtained using our object-centric data augmentation pipeline described in Sec.~\ref{subsec::data_aug}. 
Note that the test sequence and our training data are strictly separated, \ie we did not use parts of the test sequence for data augmentation.  
We trained our object coordinate network by minimizing a robust Huber loss function \cite{girshick15fastrcnn} using ADAM \cite{KingmaB14}. 
We train a separate network for each object. 
We rescale inputs and ground truth outputs for the network to 256x256px patches. 


\vspace{-0.25cm}
\subsection{Pose Estimation Accuracy}\label{subsec::sota_comparison}
\subsubsection{RGB Setup.}\label{subsec::rgb_comparison}
We estimate object poses from RGB images ignoring the depth channel. 
We evaluate the performance using the \textit{2D Projection} metric introduced by Brachmann \etal \cite{brachmann2016}. 
This metric measures the average re-projection error of 3D model vertices transformed by the ground truth pose and the estimated pose. 
A pose is accepted if the average re-projection error is less than a threshold. 

\begin{figure}
\centering
  \includegraphics[width=\linewidth]{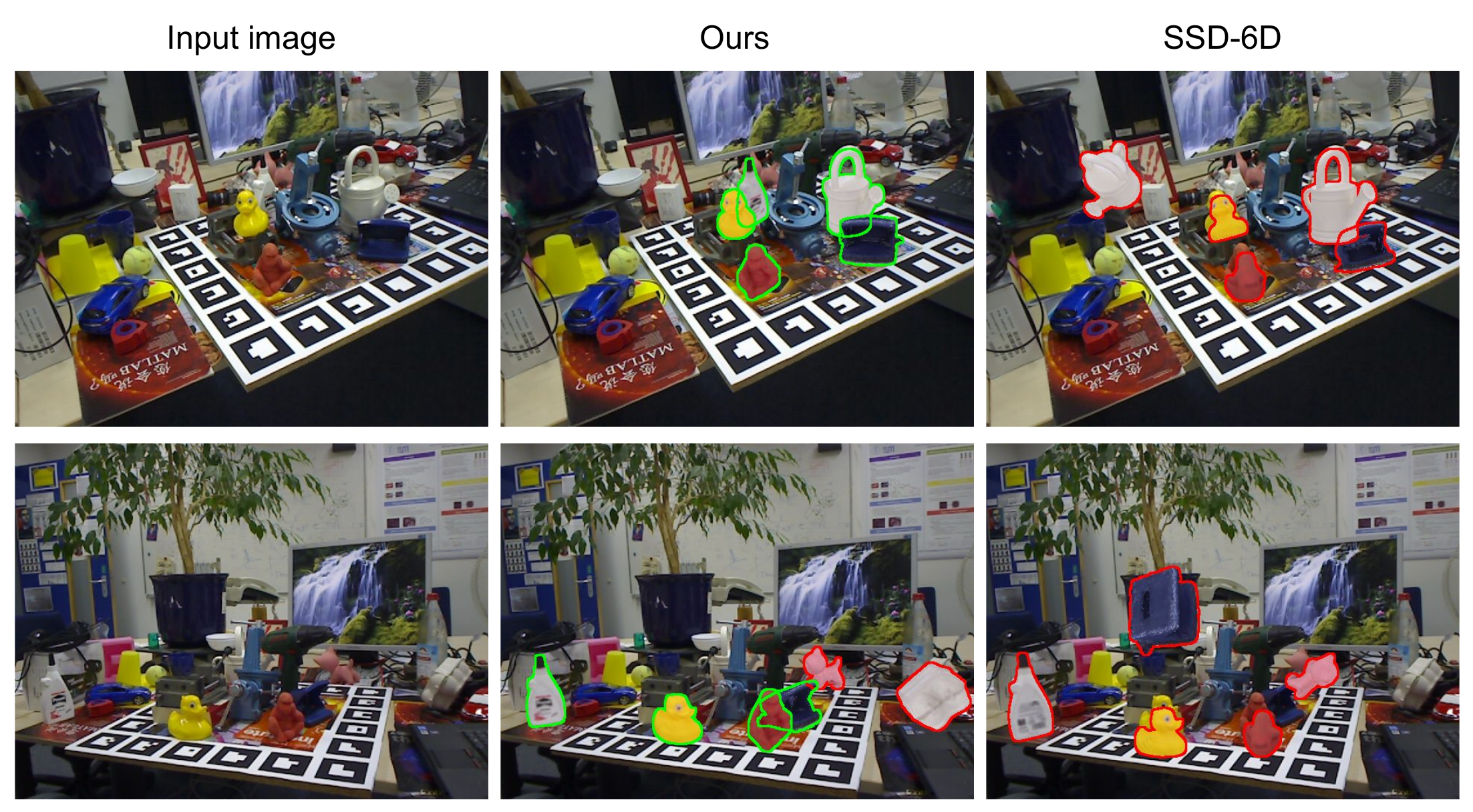}
  \caption{\textbf{Qualitative results from the RGB setup.} From left to right: input image, our results, results of SSD-6D \cite{kehl2017iccv}.
  }
  \label{fig::results_rgb}
\end{figure}

In Table \ref{tab::rgb_quantitative}, we compare the performance of our pipeline to existing RGB-based methods using two different thresholds for the 2D projection metric. 
We see that our approach outperforms the previous works for most of the objects significantly. 
Our RGB only pipeline surpasses the state-of-the-art for a 5 pixel threshold by 13\% and for a 10 pixel threshold by 39\% on average. 
Note that the results of BB8 \cite{bb82017} were obtained from image crops around the ground truth object position.
Similar to \cite{bb82017} and \cite{Tekin18cvpr}, we do not report results for \textit{EggBox} since we could not get reasonable results for this extremely occluded object using RGB only.
Note that SSD-6D \cite{kehl2017iccv} and SSS-6D \cite{Tekin18cvpr} completely fail for partly occluded objects.
We obtained the results of SSS-6D directly from \cite{Tekin18cvpr}, and of SSD-6D \cite{kehl2017iccv} using their publicly available source code and their pre-trained model. 
However, they did not release their pose refinement method, thus we report their performance without refinement. 
In the supplement, we show the accuracy of SSD-6D using different 2D re-projection thresholds.
Most of the detections of SSD-6D are far off (see also their detection performance in Fig.~\ref{tab::rf_vs_cnn}, right), therefore we do not expect refinement to improve their results much.
We show qualitative pose estimation results for the RGB setting in Fig~\ref{fig::results_rgb}.




\begin{table}
\vspace{-0.5cm}
	\centering
	\caption{\textbf{Results using RGB only.} Comparison of our pose estimation accuracy for RGB inputs with competing methods. \textit{Italic} numbers were generated using ground truth crops, thus they are not directly comparable.}
	\begin{tabular}{l|c|c|c||c|c|c|c|c}
		& \multicolumn{3}{c||}{Acceptance Threshold: 5 px} & \multicolumn{5}{c}{Acceptance Threshold: 10 px} \\
		& BB8\cite{bb82017} & Brachmann & \textbf{Ours} & BB8\cite{bb82017} & Brachmann & SSD-6D & SSS-6D & \textbf{Ours}\\
		& (GT crops) & \cite{brachmann2016} & & (\textit{GT crops}) & \cite{brachmann2016} & \cite{kehl2017iccv} & \cite{Tekin18cvpr} & \\
		\hline
		Ape  		 & \textit{28.5\%} & \textbf{31.8\%} & 24.2\%&  \textit{81.0\%} & 51.8\% & 0.5\% & 0\% & \textbf{56.1\%}\\
		Can  		 &  \textit{1.2\% }& 4.5\%  & \textbf{30.2\%} &  \textit{27.8\%} & 19.1\% & 0.6\% & 0\% & \textbf{72.4\%}\\
		Cat  		 &  \textit{9.6\%} & 1.1\%  & \textbf{12.3\%} &  \textit{61.8\%} & 7.1\%& 0.1\% & 0\% & \textbf{39.7\%}\\
		Duck 		 &  \textit{6.8\%} & 1.6\%  & \textbf{12.1\%}&  \textit{41.3\%} & 6.4\% & 0\% & 5\% & \textbf{50.1\%}\\
		Glue 		 &  \textit{4.7\%} & 0.5\%  & \textbf{25.9\%} &  \textit{37.7\%} & 6.4\% & 0\% & 0\% & \textbf{55.1\%}\\
		HoleP.       &  \textit{2.4\%} & 6.7\%  & \textbf{20.6\%}&  \textit{45.4\%} & 2.6\% & 0.3\% & 1\% & \textbf{61.2\%}\\
		\hline
		Avg 	     &  \textit{8.9\%} & 7.7\%  & \textbf{20.8\%}&  \textit{49.2\%} & 17.1\% & 0.3\% & 0.01\%& \textbf{56.0\%}\\
	\end{tabular}
	\vspace{-0.5cm}
	\label{tab::rgb_quantitative}
\end{table}

\begin{figure}
\centering
 \includegraphics[width=\linewidth]{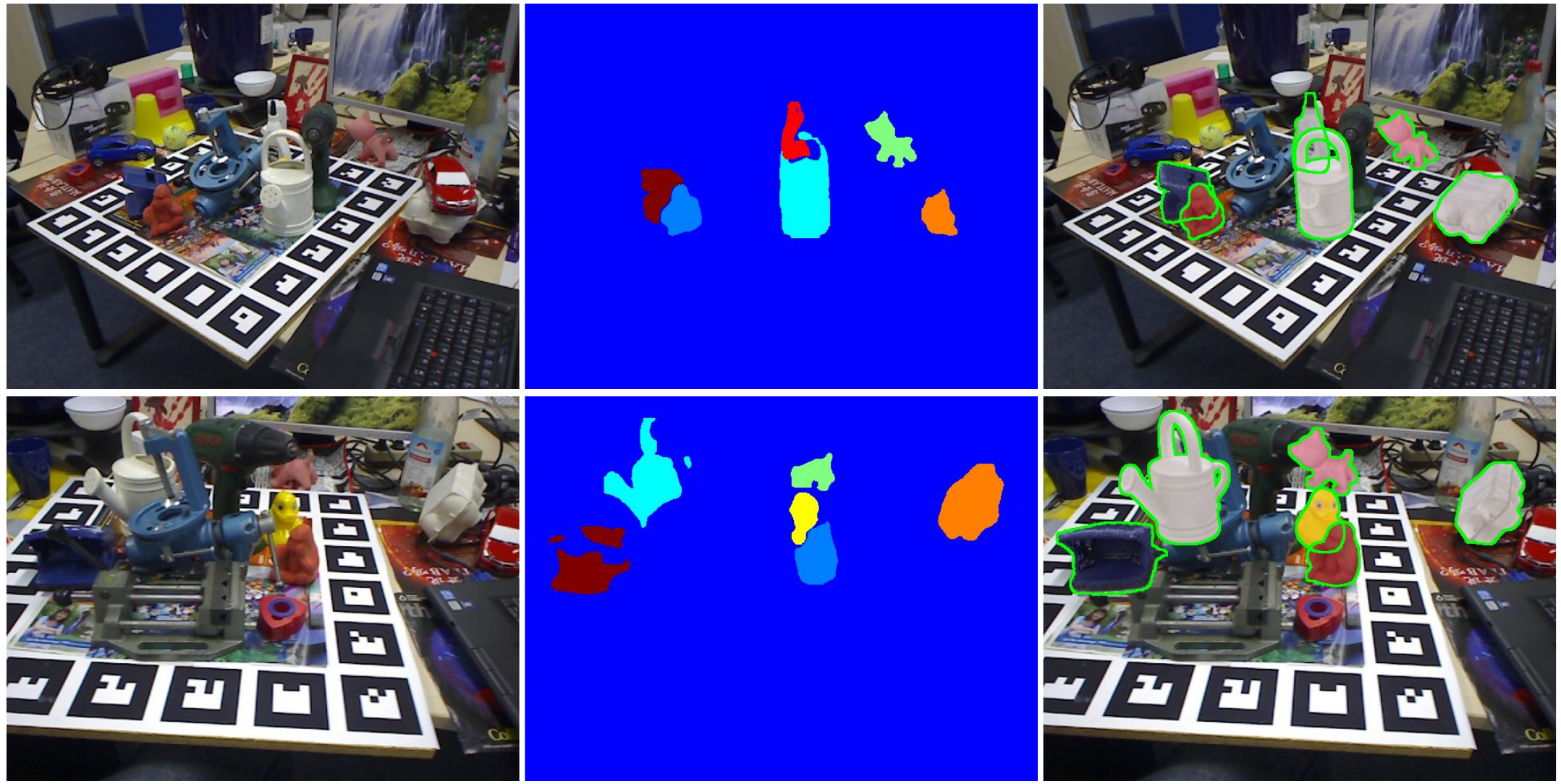}
  \caption{\textbf{Qualitative results from the RGB-D setup.} Our approach reliably estimates poses for objects which are heavily occluded. 
  The middle column shows estimated object masks of our instance segmentation step.}
  \label{fig::results_rgbd}
\end{figure}

\subsubsection{RGB-D Setup.}\label{subsec::rgbd_comparison}
\vspace{-0.25cm}
Similar to the RGB setup, we measure accuracy as the percentage of correctly estimated poses.
Following Hinterstoisser \etal \cite{Hinterstoisser12model}, we accept a pose if the average 3D distance between object model vertices transformed using ground truth pose and predicted pose lies below 10\% of the object diameter.

\begin{figure}
\centering
 \includegraphics[width=\linewidth]{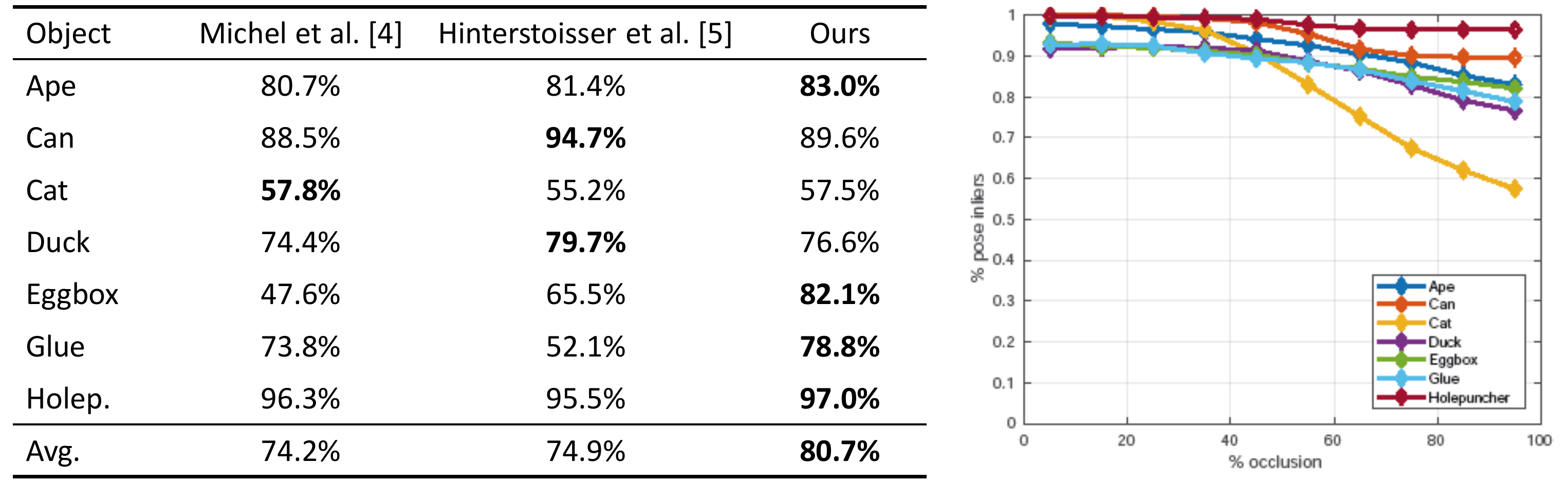}
  \caption{\textbf{Left.} Comparison of our pose estimation accuracy (RGB-D) with competing methods.
  \textbf{Right.} The percentage of correctly estimated poses as a function of the level of occlusion.}
  \label{tab::rgbd_quantitative}
\end{figure}

In Fig.~\ref{tab::rgbd_quantitative}, left, we compare the performance of our approach to Michel \etal \cite{michel2017} and  Hinterstoisser \etal \cite{Hinterstoisser2016}. 
We significantly outperform the state-of-the-art on average by 6\%, and show massive improvements for some objects. 
Fig.~\ref{fig::results_rgbd} shows qualitative results from our pipeline. 
Fig.~\ref{tab::rgbd_quantitative}, right represents the percentage of correct poses as a function of occluded object surface. 
We see that for cases of mild occlusion, our method surpasses accuracy of 90\% for all objects. 
For cases of heavy occlusion (above 60\%) our method can still recover accurate poses.


\subsubsection{Ablation Study.}\label{subsec::ablation_study}
\begin{figure}
\centering
 \includegraphics[width=\linewidth]{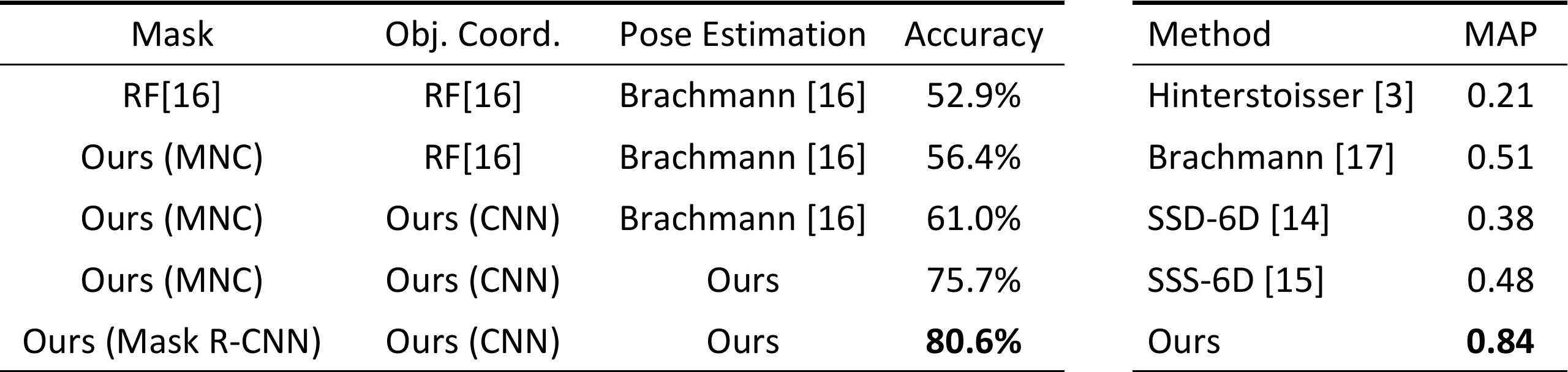}
  \caption{\textbf{Left.} Pose estimation accuracies on the RGB-D dataset using various combinations of mask estimation, object coordinates estimation and pose estimation approaches.
  \textbf{Right.} Comparison of 2D detection performance.}
  \label{tab::rf_vs_cnn}
  \vspace{-0.5cm}
\end{figure}

\vspace{-0.25cm}
We investigate the contribution of each step of our method towards the final pose estimation accuracy for the RGB-D setup. 
As discussed before, our method consists of three steps, namely instance mask estimation, object coordinate regression and pose estimation. 
We compare to the method of Brachmann \etal \cite{brachmann2014pose6d} which has similar steps, namely soft segmentation (not instance-aware), object coordinate regression, and a final RANSAC-based pose estimation.
The first two steps in \cite{brachmann2014pose6d} are implemented using a random forest, compared to two separate CNNs in our system.
Fig \ref{tab::rf_vs_cnn}, left shows the accuracy for various re-combinations of these modules. 
The first row is the standard baseline approach of \cite{brachmann2014pose6d} which achieves an average accuracy of 52.9\%. 
In the second row, we replace the soft segmentation estimated by \cite{brachmann2014pose6d} with a standard instance segmentation method, namely Multi-task Network Cascades (MNC) \cite{Dai2016CVPR}. 
The instance masks effectively constrain the 2D search space which leads to better sampling of correspondences between depth points and object coordinate predictions.
Next, we replace the object coordinate predictions of the random forest with our CNN-based predictions. 
Although we still perform the same pose optimization, this achieves an 4.6\% performance boost, showing that our encoder-decoder network architecture predicts object coordinates more precisely.
Next, we use the instance masks as above and object coordinates from our network with our geometric ICP-based refinement which further boosts the accuracy to 75.7\%. 
Finally, in the last row, we use our full pipeline with masks from Mask R-CNN followed by our other modules to achieve state-of-the-art performance of 80.7\%. 
The table clearly indicates that the accuracy of our pipeline as a whole improves when any of the modules improve, \eg by better instance segmentation.

\subsection{Instance Segmentation}\label{sub_sec::seg_eval}
 Since we cannot hope to estimate a correct pose for an object that we do not detect, the performance of instance segmentation is crucial for our overall accuracy. 
 Fig.~\ref{tab::rf_vs_cnn}, right shows the mean average precision of our method for a 2D bounding box IoU $>$ 0.5 compared to other methods. 
Since our RGB only instance segmentation network is used for both, the RGB and RGB-D setting, the MAP is equal for both settings.
 We significantly outperform all the other pose estimation methods, showing that our decoupled instance segmentation step can reliably detect objects, making the task for the following modules considerably easier.



\subsection{Object Coordinate Estimation}\label{sub_sec::objc_eval}

We trained our object coordinate network with and without our data augmentation procedure (Sec.~\ref{subsec::data_aug}). 
We measure the average inlier rate, \ie object coordinate estimates that are predicted within 2cm of ground truth object coordinates. 
When the network is trained only using the LINEMOD-M dataset, the average inlier rate is 44\% as compared to 52\% when we use the data created using our object centric data augmentation procedure. 
A clear 8\% increase in the inlier rate shows the importance of our proposed data augmentation.




\vspace{-0.2cm}
\section{Conclusion}
\vspace{-0.2cm}
We have presented \emph{iPose}, the first deep learning-based approach capable of estimating accurate poses of partly occluded objects. Our approach surpasses the state-of-the-art for both image input modalities, RGB and RGB-D. We attribute the success of our method to our decomposition philosophy, and therefore the ability to leverage state-of-the-art instance segmentation networks. We are also the first to successfully train an encoder-decoder network for dense object coordinate regression, that facilitates our robust geometric pose optimization.

\paragraph{}
\textbf{Acknowledgements:}
This work was supported by the DFG grant “COVMAP: Intelligente Karten mittels gemeinsamer GPS- und Videodatenanalyse” (RO 4804/2-1), the European Research Council (ERC) under the European Unions Horizon 2020 research and innovation programme (grant agreement No 647769) and the Heidelberg Collaboratory for Image Processing (HCI). The computations were performed on an HPC Cluster at the Center for Information Services and High Performance Computing (ZIH) at TU Dresden.

\bibliographystyle{splncs}
\bibliography{biblist}
\end{document}